\def\year{2018}\relax
\documentclass[letterpaper]{article} 
\usepackage{aaai18}  
\usepackage{times}  
\usepackage{helvet}  
\usepackage{courier}  
\usepackage{url}  
\usepackage{graphicx}  

\usepackage{booktabs}
\usepackage{multirow}
\usepackage[ruled]{algorithm2e}
\usepackage{amsmath}
\usepackage{algorithmic}
\usepackage{subcaption}

\begin{document}
%
\title{Learning Spatio-temporal Features with Partial Expression Sequences for on-the-Fly Prediction}
\author{Wissam J. Baddar \ and  Yong Man Ro\\
Image and Video Systems Lab., Electrical Engineering, \\
KAIST, South Korea\\
\{wisam.baddar,ymro\}@kaist.ac.kr\\
}

\maketitle
\begin{abstract}
Spatio-temporal feature encoding is essential for encoding facial expression dynamics in video sequences. At test time, most spatio-temporal encoding methods assume that a temporally segmented sequence is fed to a learned model, which could require the prediction to wait until the full sequence is available to an auxiliary task that performs the temporal segmentation. This causes a delay in predicting the expression. In an interactive setting, such as affective interactive agents, such delay in the prediction could not be tolerated. Therefore, training a model that can accurately predict the facial expression "on-the-fly" (as they are fed to the system) is essential. In this paper, we propose a new spatio-temporal feature learning method, which would allow prediction with partial sequences. As such, the prediction could be performed on-the-fly. The proposed method utilizes an estimated expression intensity to generate dense labels, which are used to regulate the prediction model training with a novel objective function. As results, the learned spatio-temporal features can robustly predict the expression with partial (incomplete) expression sequences, on-the-fly. Experimental results showed that the proposed method achieved higher recognition rates compared to the state-of-the-art methods on both datasets. More importantly, the results verified that the proposed method improved the prediction frames with partial expression sequence inputs.  
\end{abstract}

\section{Introduction}
Facial expressions are a channel in which humans use to non-verbally convey and communicate their internal states, emotions and intentions \cite{1}. Recently, recognizing and interpreting facial expressions have attracted researchers in computer vision, affective computing and human computer interaction fields \cite{3}. This is mainly attributed to the multitude of potential applications, such as emotionally intelligent interactive agents, fatigue measurement or even lie detection \cite{4}.

The recent success of deep learning in various computer vision tasks has influenced researchers to investigate the utilization of deep learning techniques in facial expression recognition (FER) \cite{6,7,8,9,10,11}. A FER method influenced by the concept of action units has been introduced in \cite{7,8}. Spatial features were learned from local textural patterns, namely "micro-action-patterns", using a convolutional neural network (CNN). The learned spatial features were then used to classify the facial expression \cite{7,8}. In \cite{12}, the authors proposed adopting deeper CNN models by utilizing the concept of inception layers, to improve the FER performance in expressive face images. However, those methods relied only on the spatial features, and the facial expression dynamics were not utilized, which could limit the model performance at non-apex frames or frames of subtle expression. 

Inspired by the dynamic nature of facial expressions \cite{13}, researchers investigated spatio-temporal features to capture the dynamics of facial expressions in video sequences. In \cite{9}, appearance and geometric features were learned from expressive face sequences through a fusion of a 3D CNN and a multi-layer perceptron. The authors in \cite{6}, proposed using a 3D CNN to learn spatio-temporal features from deformable facial action parts. The large number of parameters (weights and biases) in 3D CNNs that needs to be learned and the large computational complexity makes the learning processes challenging \cite{14,15} and increases the prediction time. To reduce the complexity, sampling a small number of frames or temporal normalization is usually used. This could cause a loss of information in the facial expression dynamics \cite{16}. Added to that, the aforementioned methods require the full sequence to be available at the prediction time in order to work efficiently, and cannot operate on-the-fly.

Recurrent neural networks (RNNs), in particular long short-term memory (LSTM) have been popular in encoding temporal and spatio-temporal features \cite{17}. They can provide a solid framework for on-the-fly prediction as described in \cite{18,19}. However, when training the networks in \cite{18,19}, the authors utilized dense labeled data (frame level labeling), which improves the prediction in an on-the-fly setting. On the other hand, The authors in \cite{16} utilized LSTMs to encode the spatio-temporal dynamics of facial expressions from "weakly" labeled sequences (video level labels). This improved the prediction at pre-segmented sequences but did the authors did not verify their method in an on-the-fly (interactive) scenario. Our experiments show, that relying on the weakly labeled data only, a delay in the prediction of the expression could occur (refer to Figure~\ref{fig:4}).

In this work, we proposed a method for learning spatio-temporal features for on-the-fly prediction with partial (incomplete) sequences. The proposed method facilitates the LSTMs on-the-fly prediction framework, and improves the partial prediction, such that the prediction delay problem can be resolved. To the best of our knowledge, this is the first work that estimates expression intensities, and utilizes them as dense labels to improve the expression prediction on-the-fly. As such, features can be obtained from partial (incomplete) sequences as they are fed to the model. The contributions of this paper are summarized as follows:

\begin{enumerate}
\item Obtaining densely labeled (frame level labeled) video sequences could be very costly and burdensome. Especially, in areas like FER, where labeling each frame can be subjective and time consuming [20]. On the other hand, weakly labeled data (video level labeled) is much easier to obtain. In this paper, we propose roughly generating dense labels (frame level labels) by estimating the expression intensity. The estimated expression intensities can be used in order to regulate the spatio-temporal feature training, such that prediction can be performed on-the-fly, even before the full sequence is available (using partial sequences).

\item We devise a method for learning spatio-temporal features, which can be readily used for prediction on-the-fly with partial sequences. To overcome the prediction delay problem and improve the expression prediction with partial (incomplete) expression sequences, the estimated expression intensities are used to regularize the features learning with a novel objective function construed of three objective terms. The first objective term minimizes the expression sequence classification error. The other two objective terms utilize the estimated expression intensity, and cluster the spatio-temporal feature to induce early prediction with partial sequences. As a result, the proposed method can perform prediction with partial (incomplete) sequences on-the-fly. 

\end{enumerate}

\begin{figure}[!ht]
\begin{center}
\includegraphics[width=1\textwidth,height=6.5cm,keepaspectratio]{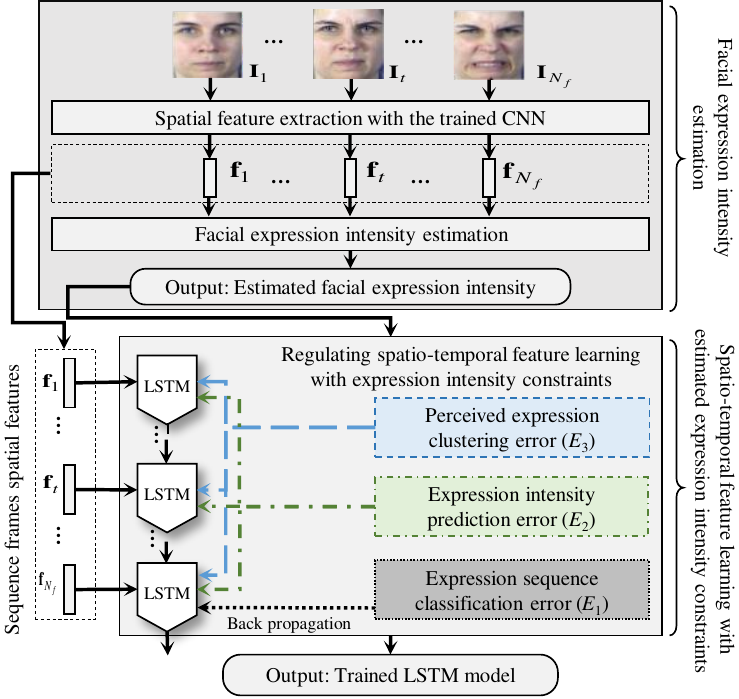}
\end{center}
   \caption{Overview of the proposed method for learning spatio-temporal features with partial expression sequences for on-the-fly prediction.}
\label{fig:1}
\end{figure}

\section{Proposed Method}
Figure 1 shows an overview of the proposed method. The proposed spatio-temporal feature learning consists of facial expression intensity estimation and feature learning with estimated expression intensity constraints. In the facial expression intensity estimation part, the trained CNN model \cite{32} is used to obtain the spatial features from each frame in the training sequences. Using the obtained features, the expression intensity of each frame is estimated with respect to the apex frame. The estimated expression intensities are used as dense labels when learning the spatio-temporal features via the LSTM. 

The spatio-temporal feature learning part focuses on encoding the facial expression dynamics, by learning spatio-temporal features via a regularized LSTM. Three objective terms are devised to guide the LSTM training. The objective terms are: (1) minimizing expression sequence classification error, (2) minimizing the expression intensity prediction error and (3) clustering spatio-temporal features obtained at each frame, by the LSTM, into perceived expression clusters. As a result, the proposed method can generate discriminative spatio-temporal features that improve the recognition of facial expression at the sequence level. Moreover, due to utilizing the expression intensity constraints, the LSTM can correctly predict the facial expressions at intermediate non-apex frames (with partial expression sequences). This allows the proposed method to be executed without requiring auxiliary tasks (such as expression transition detection, and expression segment detection \cite{23}) or delays until the prediction is valid. We detail the proposed spatio-temporal feature learning in the following subsections.
\subsection{Facial expression intensity estimation for regulating spatio-temporal feature learning}
In facial expression video sequence datasets, it is often the case that one label is given to a video sequence. However, the frames of video sequences change from a neutral expression frame to apex frame (peak expression) \cite{22}. Sometimes the sequence ends at a neutral expression frame \cite{21}. In each video sequence, the time required for expression transition (e.g., onset to apex) is different. And expression intensity varies at each frame. This makes it difficult to manually label each frame with expression intensity or give hard labels that divide the expression sequence frames. 

In this paper, we propose estimating the expression intensity at each frame with respect to the apex frame of the expression sequence. Spatial features are obtained from each cropped face region using CNN \cite{32}. Then, the cosine similarity is calculated between the spatial feature of the current frame and the apex frame by: 

\begin{equation}
cos similarity = \frac{\textup{\textbf{f}}_{t} \cdot \textup{\textbf{f}}_{apex}}{\left \|\textup{\textbf{f}}_{t}  \right \| \left \|\textup{\textbf{f}}_{apex}  \right \|},
\label{eq:1}
\end{equation}

\noindent where $\textup{\textbf{f}}_{t}$ is the spatial feature vector obtained from the frame at time $t$, $\textup{\textbf{f}}_{apex}$ is the spatial feature vector obtained from the apex frame and $\left \| \cdot  \right \|$ denotes the L2 norm operation. After the cosine similarity is obtained for all the frames of the sequence, min-max normalization is performed on the cosine similarities. As a result, the expression intensity at the apex frame is set to 1, and the expression intensity of the frame with a neutral face is set to 0. In case of a neutral expression sequence, all expression intensities are assumed to be zero.
\subsection{Spatio-temporal feature learning with estimated expression intensity constraints}
\begin{table}
  \centering
  \caption{The network architecture for learning spatio-temporal features with partial expression sequences for on-the-fly prediction. Note that $N_{f}$ is the number of frames in a sequence and $N_{c}$ is the number of expression classes. $^{*}$F and L are abbreviations for the fully connected layer and the LSTM layer, respectively. $^{**}$F$_{CNN}$ is the frame spatial feature, which is the last fully connected layer of the CNN network \cite{16}.}
  \label{Table:1}
  \resizebox{\columnwidth}{!}{
    \begin{tabular}{p{1.3cm}p{1.1cm}p{2.5cm}p{2.5cm}}
    \toprule
    \toprule
    \textbf{Type} & \textbf{Layer$^{*}$} & \textbf{Input shape} & \textbf{Output shape} \\
    \midrule
    \midrule
    Input & Input & $N_{f}\times(64\times64\times3)$ & $N_{f}\times(64\times64\times3)$ \\
    \midrule
    \multicolumn{1}{p{1.3cm}}{Spatial features}  & F$_{CNN}^{**}$ & $N_{f}\times(64\times64\times3)$  & $N_{f}\times(512\times1)$ \\
    \midrule
    \multirow{3}{1.3cm}{Spatio-\newline{}temporal\newline{}features} & L${1}$ & $N_{f}\times(512\times1)$ & {$N_{f}\times(512\times1)$} \\
    
    & {F${f}$} & {$N_{f}\times(512\times1)$} & {$N_{f}\times1$} \\
    
    & {F${c}$} & {$N_{f}\times(512\times1)$} & {$N_{c}\times1$} \\
   
    \midrule
    \midrule
    \end{tabular}%
    }
\end{table}%
Table~\ref{Table:1} summarizes the overall architecture of the proposed network. To learn the spatio-temporal features with estimated expression intensity constraints, the spatial features are extracted from the sequence frames via the learned CNN model and fed to the LSTM. An objective function with three objective terms is devised to learn the LSTM network parameters.
 
The first objective term of the objective function ($E_1$) is devised for minimizing the expression sequence classification error at the expression output layer of the network (F$_c$ in Table~\ref{Table:1}). To that end, a cross-entropy error \cite{24} is utilized to enforce learning a discriminative feature, which is defined as:
 
\begin{equation}
E_{1}= -\sum_{i,c}y_{i,c} \textup{log}\widehat{y}_{i,c}^{(F_c)},
\label{eq:2}
\end{equation}

\noindent where $c$ is the expression class index,$y_{i,c}$ is the expression class ground truth of the $i\textup{-th}$ sequence (1 if $c$ is the correct class and 0 otherwise), and $\widehat{y}_{i,c}^{(F_c)}$ is the predicted probability that the sequence belongs to the class $c$ calculated at the output layer (F$_c$ in Table~\ref{Table:1}).

To perform back-propagation, the gradients of the objective term $E_1$ for the layer F$_c$ are written as follows:
 
\begin{equation}
\begin{split}
\frac{\partial E_1}{\partial \textup{\textbf{b}}^{(F_c)}}=\sum_{i}\left ( \widehat{y}_{i}^{(F_c)}-y_{i}^{(F_c)} \right ) \\ \textup{and}  \frac{\partial E_1}{\partial \textup{\textbf{W}}^{(F_c)}}=\textup{\textbf{h}}^{(L_1)} \left ( \frac{\partial E_1}{\partial \textup{\textbf{b}}^{(F_c)}} \right ) ^{T}, 
 \end{split}
\label{eq:3}
\end{equation}

\noindent where $\textup{\textbf{b}}^{(F_c)}$ and $\textup{\textbf{W}}^{(F_c)}$ denote the biases and weights of the layer F$_c$, respectively.$\textup{\textbf{h}}^{(L_1)}$ denotes the output of the LSTM layer (L$_\textup{1}$ in Table~\ref{Table:1}), $\textbf{y}_i = [{y}_{i,1},{y}_{i,2},...,{y}_{i,c},...{y}_{i,N_c}]^T$ is the expression ground truth vector of the $i-th$ training sample,${y}_{i,c}$ if and only if the true class of the sample is $c$, and $\widehat{y}_{i}^{(F_c)}$ denotes the predicted probability calculated at the layer F$_c$ with a Softmax function. As shown in ~\ref{eq:3}, the gradient with respect to the weight is dependent on the gradient with respect to the bias. Therefore, the gradients with respect to the bias are sufficient to be described in the objective term. The gradients of the other layers can also be computed by the same back-propagation algorithm \cite{25}. 

The cross-entropy loss of the objective term for minimizing the expression sequence classification error ($E_1$) is calculated at the end of a sequence. As a result, the optimization highly affects the spatio-temporal feature (L$_1$ output from Table~\ref{Table:1}) at the last frame of sequence ($N_f\textup{-th}$ frame). However, the expression intensity variations of expression sequences result in different spatio-temporal features ($_1$  output) at each frame. Hence, miss-classification could occur when prediction is performed at early stages of the sequence (at non-apex frames with a partial expression sequence input). The prediction of the correct expression could be delayed until the end of the sequence (apex, or near apex frames). To mitigate the aforementioned delay problem, we propose the second objective term ($E_2$), which minimizes expression intensity prediction error. To that end, Euclidian loss is used to enforce learning discriminative features at each frame, which is defined as:

\begin{equation}
E_{2}= \frac{1}{2}\sum_{i,c}\left \| \mathbf{\widehat{L}}_{c,i}^{(F_f)} -\mathbf{L}_{c,i} \right \|_{2}^{2},
\label{eq:4}
\end{equation}

\noindent where $\mathbf{L}_{c,i}$ is a vector of expression intensity labels for the $i-th$ sequence of the $c-th$ class (calculated by the estimation method described in section 2.1), and $\mathbf{\widehat{L}}_{c,i}^{F_f}$ is a vector of the predicted expression intensities at each frame.
 
The gradient of the second objective term ($E_2$) in ~\ref{eq:4}, with respect to the bias of the last LSTM layer (L$_1$ in Table~\ref{Table:1}) can be written as:
 
\begin{equation}
\begin{split}
\frac{\partial E_2}{\partial \textup{\textbf{b}}^{(L_1)}}=\sum_{c,p,i}g'(e_2)\left ( \widehat{\textup{\textbf{L}}}_{c,i}^{(F_f)}-\textup{\textbf{L}}_{c,i} \right ) \circ \sigma'(\textup{\textbf{h}}^{(L_1)}),\\ 
\textup{where } e_{2}= \left \| \mathbf{L}_{c,i}- \mathbf{\widehat{L}}_{c,i}^{(F_f)} \right \|_{2}^{2},
\end{split}
\label{eq:5}
\end{equation}

\noindent where '$\circ$' denotes the Hadamard product,$\textup{\textbf{b}}^{(L_1)}$ is the bias of the LSTM layer (L$_1$ in Table ~\ref{Table:1}), $\sigma(.)$ is the activation function, and $\textup{\textbf{h}}^{(L_1)}=\textup{\textbf{W}}_o^{(l)}[\textup{\textbf{h}}_{t-1}^{(l)}, \textup{\textbf{h}}_t^{(l-1)}] +  \textup{\textbf{b}}_o^{(l)}]$ \cite{27}. 

\begin{figure}
\begin{center}
\includegraphics[width=0.9\linewidth] {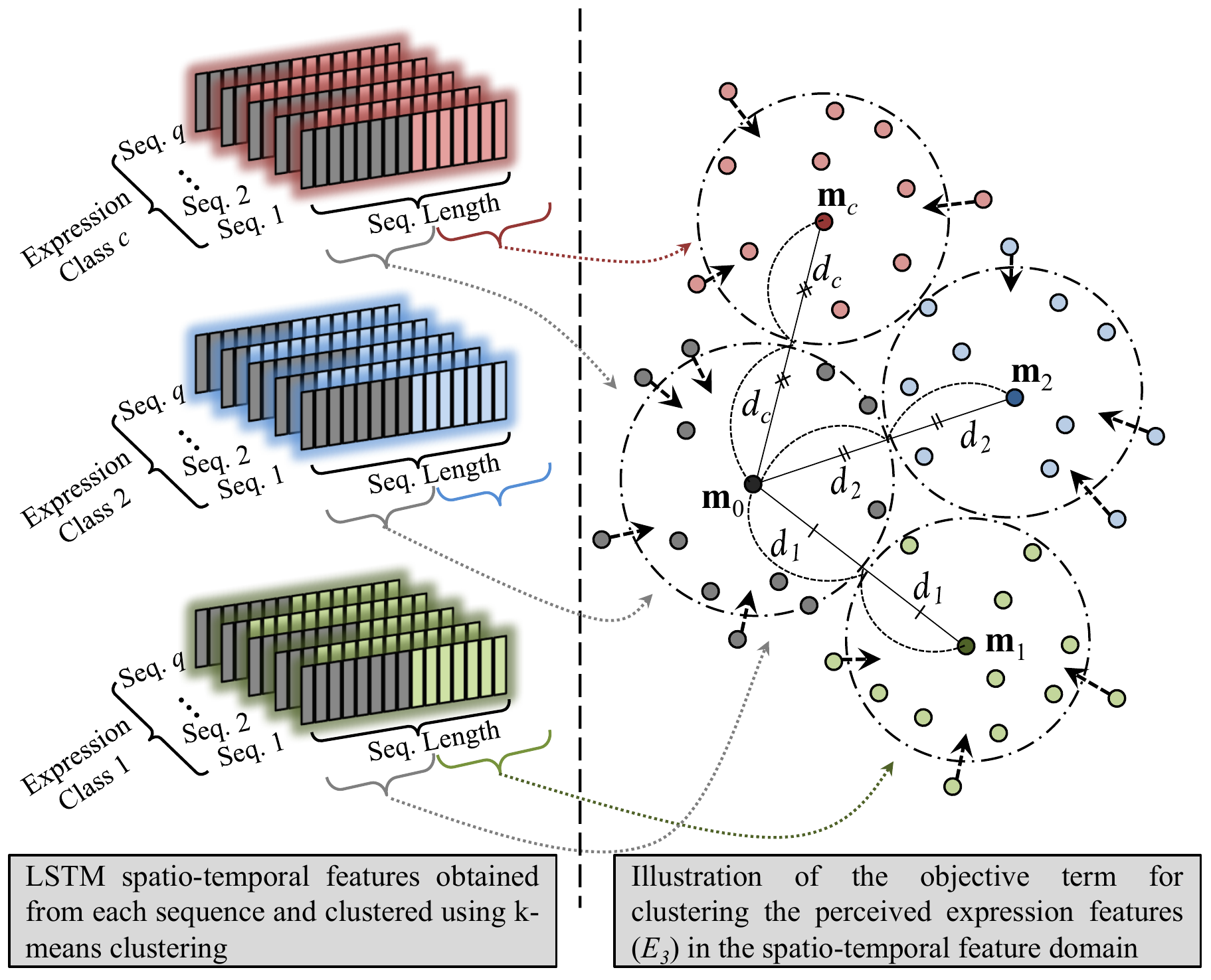}
\end{center}
   \caption{Illustration of the objective term for clustering the spatio-temporal features (L$_1$ output from Table~\ref{Table:1}) obtained at each frame. The features are clustered into perceived expression clusters (i.e., frames perceived as neutral and frames perceived as the expression class). The left part of the figure illustrates the LSTM features obtained at each frame in expression sequences. The features are clustered (using the k-means clustering) as a perceived expressive class. The features clustered as perceived expression classes are shown in color (red, blue and green). The features clustered as neutral (perceived neutral frames) are shown in dark gray. The right side of the figure illustrates the perceived expression clusters in the feature space. Each dot represents the perceived expression class. The circle enclosing the colored dots represents the cluster of a certain expression class. (best viewed in color)}
\label{fig:2}
\end{figure}

When expression sequences transition from a neutral state to the apex. Humans would perceive the first number of frames as neutral frames, and start perceiving the expression at later stage. Hence, the learned spatio-temporal features are supposed to predict neutral expression at the beginning of the sequence, and predict the expression at later frames. To improve the perceived expression prediction at early stages of the expression sequence (prediction with partial expression sequence inputs) a third objective term ($E_3$) is devised. For each expression, the spatio-temporal features from the LSTM (L$_1$ output in Table~\ref{Table:1}) are obtained from all frames of all the expression sequences of that class. Then, the features for certain class (e.g., feature from smile sequences) are clustered into 2 classes, namely, perceived neutral and perceived expression class (e.g., smile, disgust, etc.). K-means clustering \cite{28} is utilized for the clustering. Figure~\ref{fig:2} (left) illustrates an example of sequences after k-means clustering. As shown in the figure, the number of frames perceived as neutral or expressive class, can be different from sequence to sequence. However, because the k-means cluttering is non-supervised, the perceived expressive frames features and the perceived neutral frames features can be obtained automatically. Note that in the case of neutral sequences, all frames are assigned to the perceived neutral expression. 

To induce early prediction at non-apex frames (when the prediction is performed with partial expression sequence inputs), $E_3$ clusters the spatio-temporal features, obtained from the sequence frames, into perceived expression clusters (i.e., frames perceived as neutral and frames perceived as the expression class). It minimizes the intra-class variation due to expression intensity variations or subject appearance variations.  The objective term $E_3$ is illustrated in Figure~\ref{fig:2} (right). It can be written as:
 
\begin{equation}
E_{3}= \frac{1}{2}\sum_{c,p,i} g\left ( \left \| \mathbf{f}_{c,p,i} -\mathbf{m}_{c} \right \|_{2}^{2} - d_c^2 \right ),
\label{eq:6}
\end{equation}

\noindent where $\mathbf{f}_{c,p,i}$ is the spatio-temporal feature of the $i\textup{-th}$ expression sequence of class $c$ and obtained from LSTM layer (L1 in Table 1) at the $p\textup{-th}$ frame of the sequence. $\mathbf{m}_{c}$ is the mean feature vector of spatio-temporal features with the perceived class $c$. $d_c$ is half the distance between $\mathbf{m}_{c}$ and $\mathbf{m}_{0}$, where $\mathbf{m}_{0}$ is the mean of the perceived neutral frames. The function $g(\omega)$ is a smoothed approximation of $[\omega]_+=\textup{max}(0,\omega)$ defined as \cite{29}:

\begin{equation}
g(\omega )= \frac{1}{\beta}log\left ( 1+ \textup{exp}(\beta\omega)\right ),
\label{eq:7}
\end{equation}

\noindent where $\beta$ is a sharpness parameter. 

For simplicity when obtaining the gradient of the third objective term ($E_3$) in (~\ref{eq:7}), the mean feature vector $\mathbf{m}_{c}$ is assumed to be a constant vector. Thus the gradient of $E_3$ with respect to the bias of the LSTM layer (L$_1$ in Table~\ref{Table:1}) can be written as:
 
\begin{equation}
\begin{split}
\frac{\partial E_3}{\partial \textup{\textbf{b}}^{(L_1)}}=\sum_{c,p,i}g'(e_3)\left ( \textup{\textbf{f}}_{c,p,i}-\textup{\textbf{m}}_{c} \right ) \circ \sigma'(\textup{\textbf{h}}^{(L_1)}), \\
\textup{where } e_{3}= \left \| \mathbf{m}_{c}- \mathbf{f}_{c,p,i} \right \|_{2}^{2} - (d_c)^2.
\end{split}
\label{eq:8}
\end{equation}
\subsection{Training the LSTM with the proposed objective terms}
The training process was performed in three steps detailed in Algorithm~\ref{Alg:1}. The training starts with the spatial features obtained from each frame in the training sequences, expression sequence labels and the estimated expression intensities. For each epoch, the network is first trained with the objective terms $E_1$ and $E_2$. In this paper, we assumed that both objective terms contribute equally to updating the LSTM weights (i.e., each loss was given the same weight). In the second training step, features are obtained from all frames of all the sequence of the training set. K-means clustering is then utilized to obtain the mean of perceived expressions in each expression class (see Figure 2). Finally, the obtained mean and distance values are fixed and utilized to update the LSTM weights according to objective term $E_3$.
\begin{algorithm}[!ht]
\caption{Pseudo code for training the LSTM with the proposed objective terms}
\label{Alg:1}
\algsetup{linenosize=\tiny}
\scriptsize
\textbf{Input: X:} Spatial features from expression sequences \\
\quad \quad \quad \textbf{Y:}the facial expression sequences labels \\
\quad\quad\quad \textbf{L:} the estimated expression intensities \\

 \textbf{Output:} Trained LSTM model 
 
 LSTM parameter initialization\;
 \While{epoch  \textless $N_{epochs}$}{
 \While{batch  \textless $M_{batches}$}{
  Feedforward (\textbf{X}(batch),\textbf{Y}(batch),\textbf{L}(batch))\;
  Get gradient for $E_1+E_2$ \& update weights
  }
  
  \For{expression class in (1:C)}{
        \textbf{f}$_c$ = Get class sequences features\;
        }

  \For{expression class in (1:C)}{
        \textbf{m}$_c$ = K-means(\textbf{f}$_c$ , K = 2)\;
        \textbf{d}$_c$ = Calculate distances\;
        }
          
 \While{batch  \textless $M_{batches}$}{
  Feedforward (\textbf{X}(batch),\textbf{m}$_c$(batch),\textbf{d}$_c$(batch))\;
  Get gradient for $E_3$ \& update weights
  }
 }
\end{algorithm}
\section{Experiments}
\subsection{Experimental setup}
To verify the effectiveness of the proposed method, experiments were conducted on two datasets. The face region was detected and facial landmark detection was performed \cite{30} on each frame. The face region was automatically cropped and aligned based on the eye landmarks \cite{31}. The construction of the utilized MMI and Oulu-CASIA datasets was performed as follows: 

\begin{enumerate}

\item MMI dataset \cite{21}: A total of 205 expression sequences with frontal faces were collected from 30 subjects. Each expression sequence was labeled with one of the six basic expressions classes (i.e., angry, disgust, fear, happy, sad, and surprise). In the MMI dataset, an expression sequence was recorded from onset to apex to offset. The indexes of the apex frames were located manually \cite{32}. To unify the experiments on both datasets (MMI and Oulu CASIA), only onset to apex frames were used. For a practical FER system, neutral sequences are required to be included during the training. To that end, neutral sequences of the subject were simulated by manually collecting frames from the beginning and the end of the expressive sequences. As such, sequences of cropped neutral expression faces were generated. As total of 38 natural expression sequences were generated and added the total to the MMI sequences.
 
\item Oulu-CASIA dataset \cite{22}: Sequences of the six basic expressions were collected from 80 subjects under three illumination conditions. For the experiments, a total of 480 image sequences were collected from sequences captured with a visible light camera under normal illumination conditions. For each subject, the basic expression sequence was captured from a neutral face until the expression apex. Similar to the MMI dataset, 80 neutral expression sequences were simulated by manually collecting natural frames from different sequences of the same subject.
\end{enumerate}

All the experiments in this paper were conducted in a subject independent manner, such that the subjects in the training set were excluded from the test set. In particular, experiments on the MMI dataset set were performed with a leave-one-subject-out (LOSO) cross validation \cite{3,6,16,33}, while 10-fold cross validation \cite{9} was used for the experiments conducted on the Oulu-CASIA dataset. 

To avoid overfitting, due to the limited number of samples in the utilized datasets (MMI and Oulu-CASIA), data augmentation was performed during the network training \cite{10,16}. For the CNN training, 54 augmentation variations of each expressive image were obtained by: (1) horizontal flipping of the sequence frames, (2) rotating the frames between the angles $[-5^\circ , 5^\circ]$ with an increment of $1^\circ$, (3) translating the frames along $[\pm3, \pm3]$ pixels in the x and y axis with 1 pixel increments, and (4) scaling the frames with scaling factors of 0.90, 0.95, 1.05 and 1.10.

For training the LSTM, each frame in the sequence was augmented similar to the aforementioned augmentation processes for training the CNN. From each augmented sequence two temporal augmentations were performed by selecting even and odd frames from each sequence. As a result, 108 augmentations of the expression sequences were generated and used to train the LSTM network. 

The learning and implementation of the CNN and LSTM network (shown in Table~\ref{Table:1}) was done using TensorFlow. For the activation function, rectified linear unit (ReLU) was used in all layers except the layer FCNN, in which sigmoid activation was utilized in order to bound the LSTM input features and insure the LSTM learning stability. In this paper, CNN initial learning rate was set to 0.0001 for both the MMI and the Oulu-CASIA datasets, and the training was performed for 30 epochs. For the LSTM, the learning rate was set to 0.0001, and the learning rate was reduced by a factor of 10 every 10 epochs. The LSTM training was conducted for 50 epochs.
\subsection{Effectiveness of the proposed method compared with previous state-of-the-art and existing methods on sequence-level prediction}
To demonstrate the effectiveness of the proposed method, the FER performance of the proposed method was compared to previously reported state-of-the-art and existing methods \cite{6,7,8,9}. The experiment was conducted under LOSO cross validation. From previously reported methods, static (frame-based) features (i.e., AURF \cite{7} and AUDN \cite{8}) were evaluated at the apex frames. Spatio-temporal feature based methods (i.e., 3D CNN-DAP \cite{6}, DTAGN \cite{9} and CNN+LSTM \cite{16}) were evaluated at the sequence level. The comparative recognition rates are shown in Table~\ref{Table:2}. As shown in the table, the proposed method outperformed existing state-of-the-art FER methods. Specifically, the proposed method showed better recognition rates compared to the deep learning based methods with spatial features (AURF and AUDN). This can be attributed to the efficient encoding of the expression dynamics in the proposed spatio-temporal features learning. Moreover, by utilizing the estimated expression intensity in the proposed objective function, the proposed method achieved a superior performance to methods that encoded the spatio-temporal dynamics with a 3D CNN (3D CNN-DAP, DTAGN and CNN+LSTM).

Table~\ref{Table:3} shows the recognition rate comparisons with state-of-the-art spatio-temporal feature based methods on the Oulu-CASIA dataset. The experiment was conducted with 10-fold subject-independent cross validation. As shown in the table, the results indicate that the proposed method is more effective than previous state-of-the-art spatio-temporal feature based methods.

\begin{table}
  \centering
  \caption{FER performance comparisons with existing FER methods on the MMI dataset in terms of recognition rate.*6 expression classes represent the basic expressions (i.e., angry, disgust, fear, happy, sad, and surprise).*7 expression classes represent the basic expressions + neutral}
  \label{Table:2}
  \resizebox{\columnwidth}{!}{
    \begin{tabular}{p{4.5cm}p{1.8cm}p{0.9cm}p{1.8cm}}
    \toprule
    \toprule
    \multicolumn{1}{c}{\textbf{Method}} & \multicolumn{1}{p{1.8cm}}{\textbf{Expression\newline{}Classes*}} & \multicolumn{1}{p{0.9cm}}{\textbf{Input}} & \multicolumn{1}{p{1.8cm}}{\textbf{Recognition rate(\%)}} \\
   
    \midrule
    \multicolumn{1}{p{4.5cm}}{3D CNN-DAP \cite{6}} & \multicolumn{1}{c}{6} & \multicolumn{1}{c}{sequence} & \multicolumn{1}{c}{63.4} \\
    \multicolumn{1}{p{4.5cm}}{DTAGN \cite{9}} & \multicolumn{1}{c}{6} & \multicolumn{1}{c}{sequence} & \multicolumn{1}{c}{70.24} \\
    \multicolumn{1}{p{4.5cm}}{AURF \cite{7}} & \multicolumn{1}{c}{7} & \multicolumn{1}{c}{static} & \multicolumn{1}{c}{69.88} \\
    \multicolumn{1}{p{4.5cm}}{AUDN \cite{8}} & \multicolumn{1}{c}{7} & \multicolumn{1}{c}{static} & \multicolumn{1}{c}{75.85} \\
    \multicolumn{1}{p{4.5cm}}{CNN+LSTM \cite{16}}  & \multicolumn{1}{c}{6} & \multicolumn{1}{c}{sequence} & \multicolumn{1}{c}{78.61} \\
    \multicolumn{1}{p{4.5cm}}{CNN+LSTM \cite{16}}  & \multicolumn{1}{c}{7} & \multicolumn{1}{c}{sequence} & \multicolumn{1}{c}{76.64} \\
    \multicolumn{1}{p{4.5cm}}{Proposed method($E_1+ E_2$)} & \multicolumn{1}{c}{7} & \multicolumn{1}{c}{sequence} & \multicolumn{1}{c}{\textbf{78.29}} \\
    \multicolumn{1}{p{4.5cm}}{Proposed method($E_1+E_2+E_3)$} & \multicolumn{1}{c}{7} & \multicolumn{1}{c}{sequence} & \multicolumn{1}{c}{\textbf{78.96}} \\
    \midrule
    \midrule

    \end{tabular}%
    }
\end{table}%

\begin{table}
  \centering
  \caption{FER performance comparisons with existing FER methods on the Oulu-CASIA dataset in terms of recognition rate. Note that all these methods use sequences as input.}
  \label{Table:3}
  \resizebox{\columnwidth}{!}{
    \begin{tabular}{p{5.5cm}p{1.8cm}p{1.8cm}}
    \toprule
    \toprule
    \multicolumn{1}{c}{\textbf{Method}} & \multicolumn{1}{p{1.8cm}}{\textbf{Expression\newline{}Classes*}}  & \multicolumn{1}{p{1.8cm}}{\textbf{Recognition rate(\%)}} \\
    \midrule
    \midrule

    \multicolumn{1}{p{7cm}}{LBP-TOP \cite{36}} 				& \multicolumn{1}{c}{6}     & \multicolumn{1}{c}{68.13} \\
    \multicolumn{1}{p{7cm}}{HOG 3D \cite{37}}  				& \multicolumn{1}{c}{6}     & \multicolumn{1}{c}{70.63} \\
    \multicolumn{1}{p{7cm}}{AdaLBP \cite{22}}  				& \multicolumn{1}{c}{6}     & \multicolumn{1}{c}{73.54} \\
	\multicolumn{1}{p{7cm}}{Atlases \cite{38}}  				& \multicolumn{1}{c}{6}     & \multicolumn{1}{c}{75.52} \\
    \multicolumn{1}{p{7cm}}{Dis-ExpLet \cite{39}} 			& \multicolumn{1}{c}{6}     & \multicolumn{1}{c}{76.65} \\

    \midrule

	\multicolumn{1}{p{7cm}}{DTAGN \cite{9}} 					& \multicolumn{1}{c}{6}     & \multicolumn{1}{c}{81.46} \\
    \multicolumn{1}{p{7cm}}{CNN+LSTM \cite{16}}  				& \multicolumn{1}{c}{7}     & \multicolumn{1}{c}{78.21} \\
    \multicolumn{1}{p{7cm}}{Proposed method($E_1+E_2$)} 		& \multicolumn{1}{c}{7}     & \multicolumn{1}{c}{\textbf{82.15}} \\
    \multicolumn{1}{p{7cm}}{Proposed method($E_1+E_2+E_3$)} 	& \multicolumn{1}{c}{7}     & \multicolumn{1}{c}{\textbf{82.86}} \\
    
    \bottomrule
    \bottomrule
    \end{tabular}%
}
\end{table}%

\begin{figure}
\begin{center}
\includegraphics[width=0.7\linewidth,height=3.8cm] {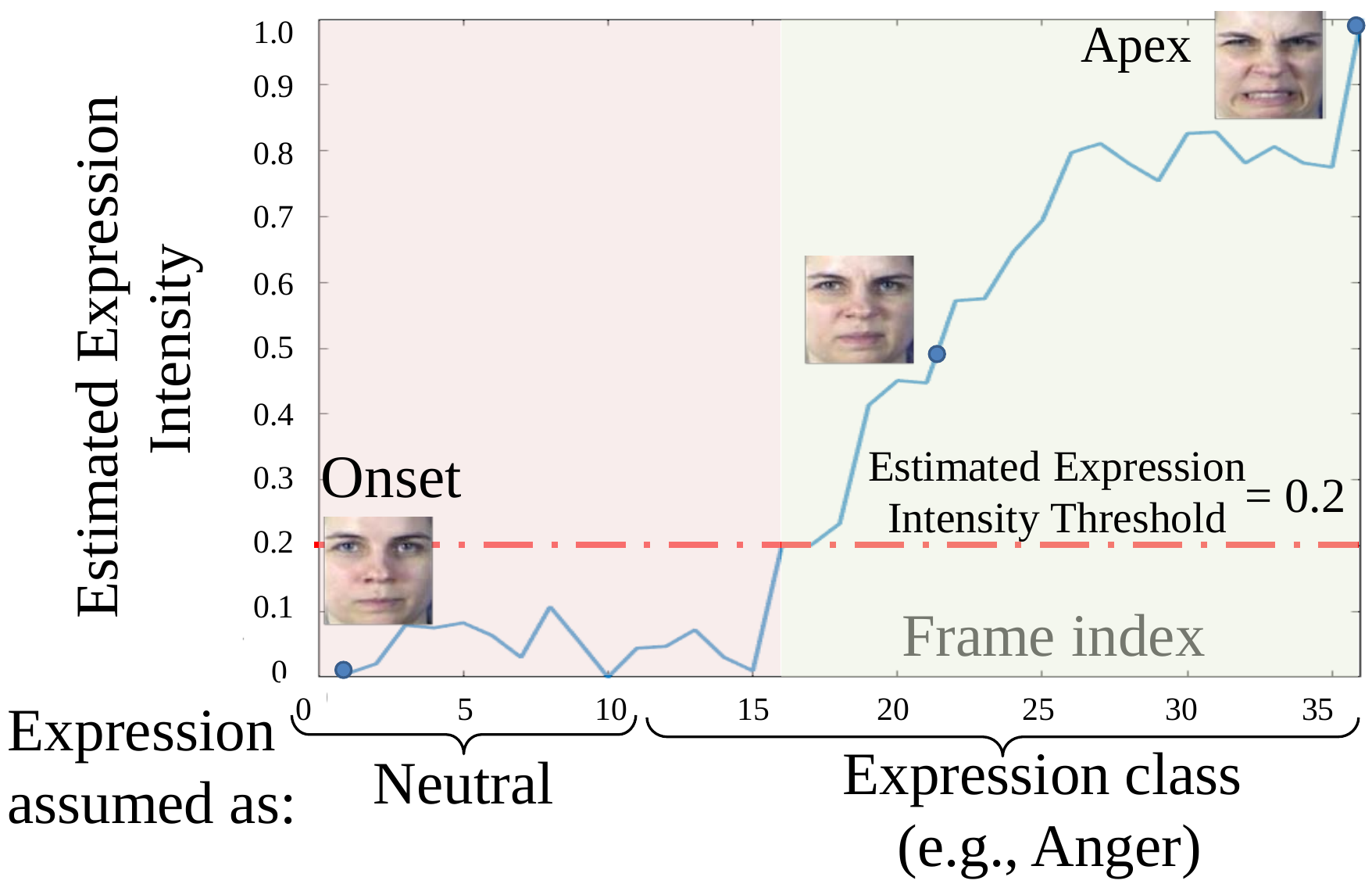}
\end{center}
   \caption{Generating frame level (dense) labels at different expression intensities to validate the FER performance at different facial expression intensities.}

\label{fig:3}
\end{figure}

\begin{figure}
\begin{center}
\includegraphics[width=1\linewidth] {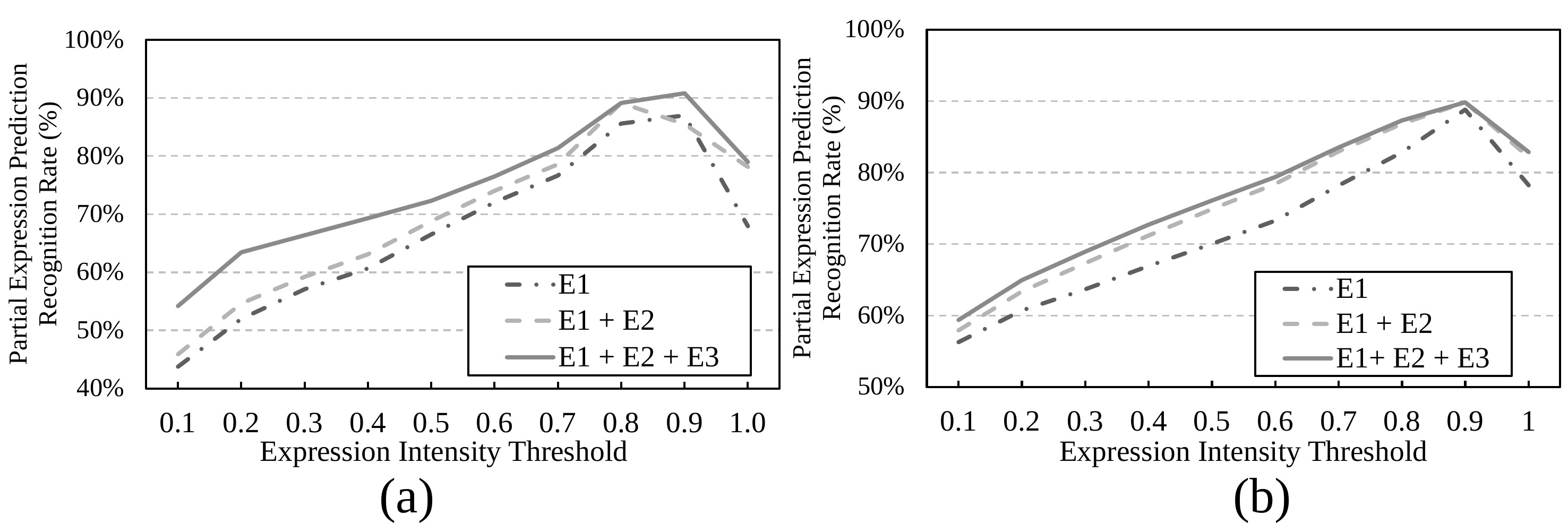}
\end{center}
   \caption{FER performance at different expression intensity thresholds. (a) MMI Dataset (b) Oulu Dataset.}
\label{fig:4}
\end{figure}

\begin{figure*}[!ht]
    \centering
    
    \begin{subfigure}[b]{0.3\textwidth}
                 \centering
                 \includegraphics[width=1\textwidth,height=2.5cm,keepaspectratio]{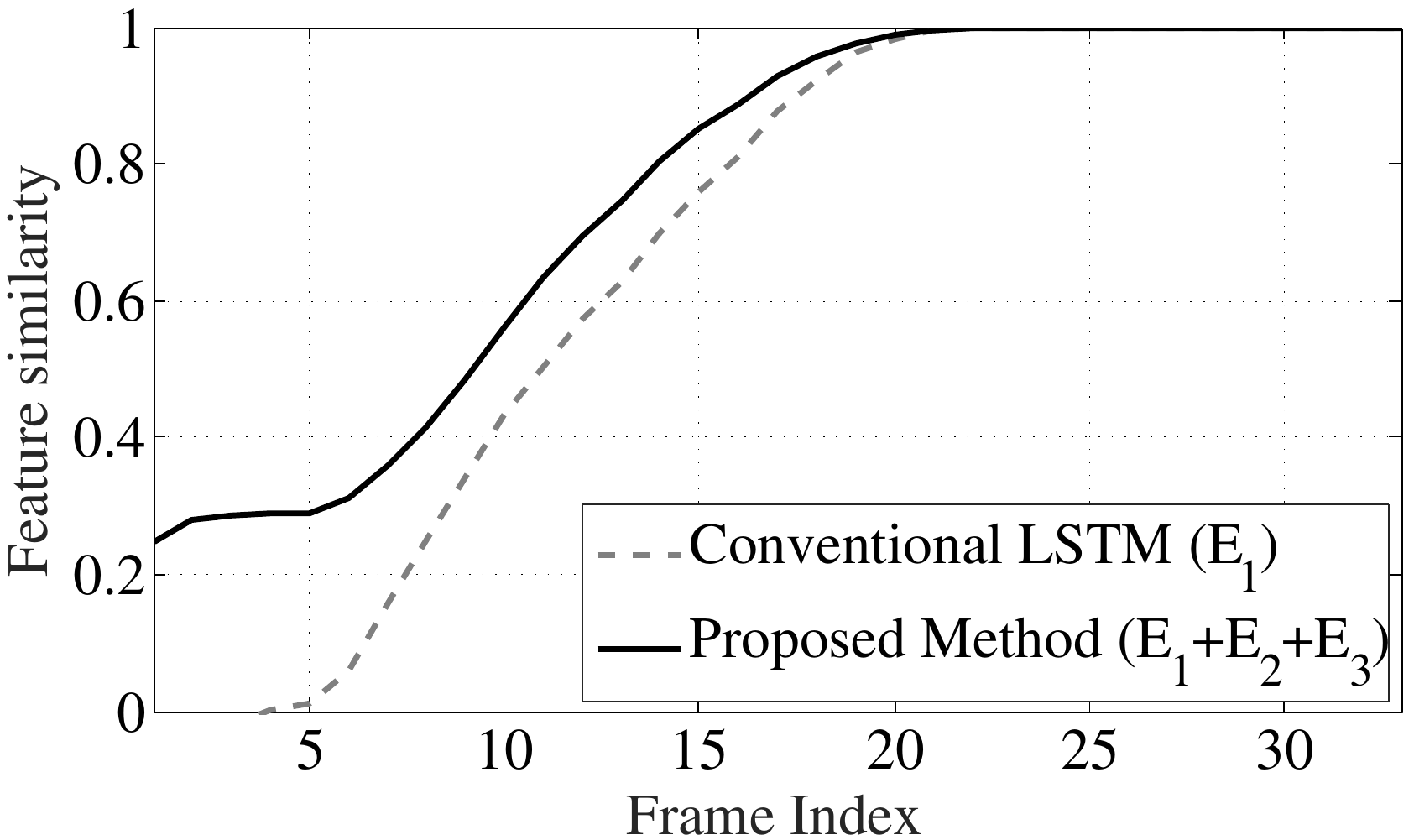}
                 \caption{}
             	 \label{fig5:a}
         \end{subfigure}
         \begin{subfigure}[b]{0.3\textwidth}
                 \centering
                 \includegraphics[width=1\textwidth,height=2.5cm,keepaspectratio]{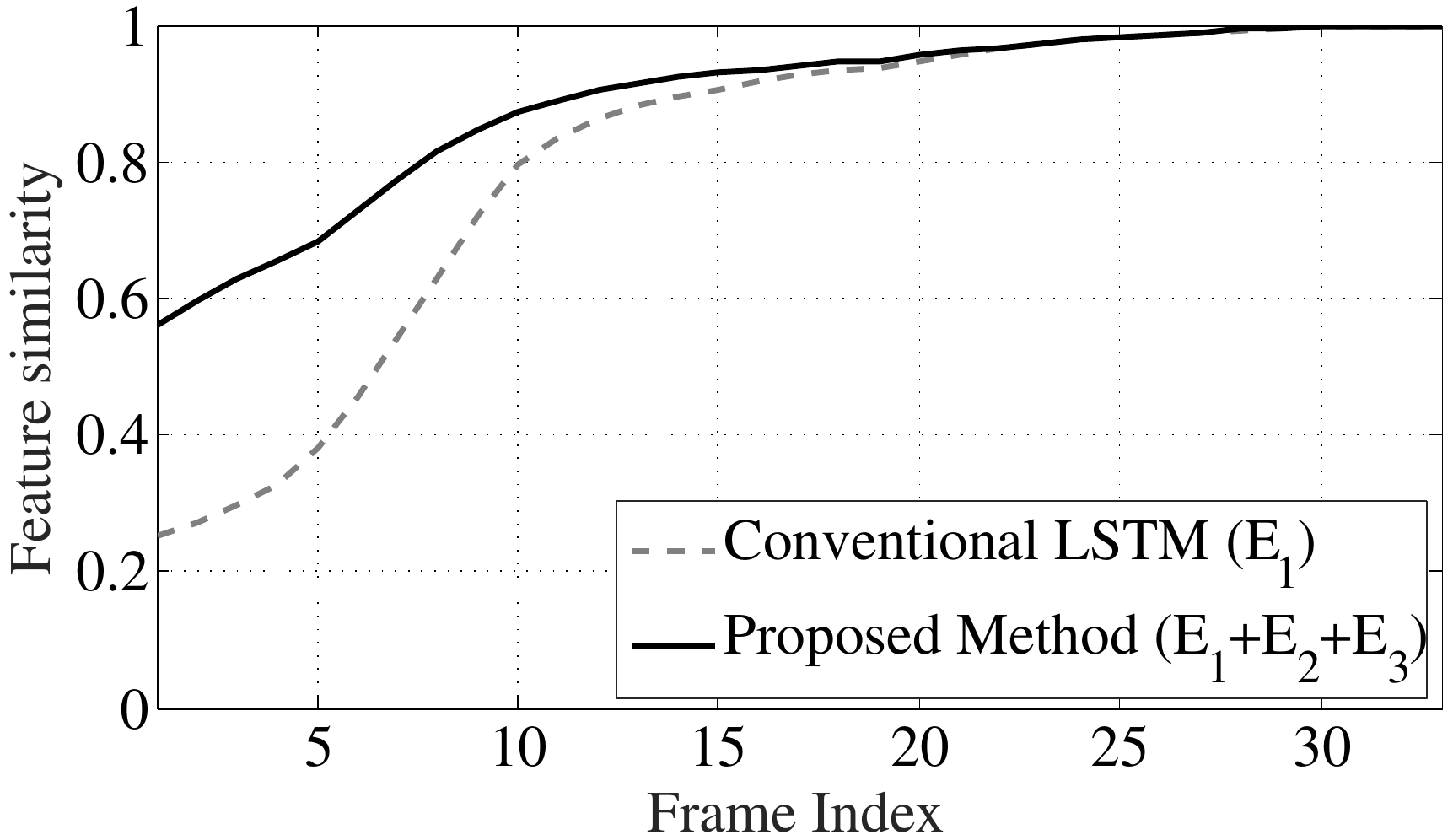}
                 \caption{}
             	 \label{fig5:b}
         \end{subfigure}%
         \begin{subfigure}[b]{0.3\textwidth}
                 \centering
                 \includegraphics[width=1\textwidth,height=2.5cm,keepaspectratio]{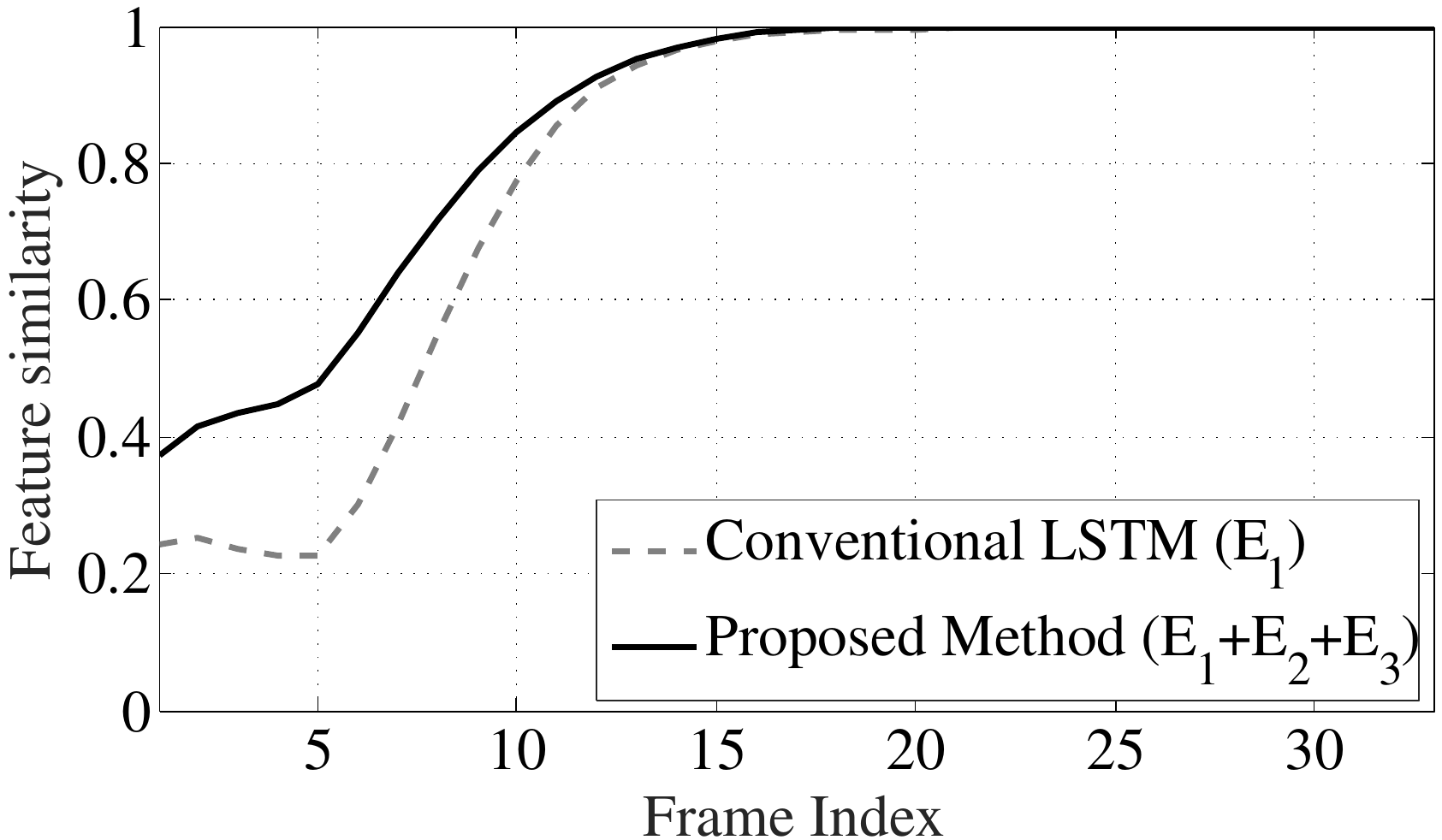}
                 \caption{}
             	 \label{fig5:c}
         \end{subfigure}

     \begin{subfigure}[b]{0.3\textwidth}
             \centering
             \includegraphics[width=1\textwidth,height=2.5cm,keepaspectratio]{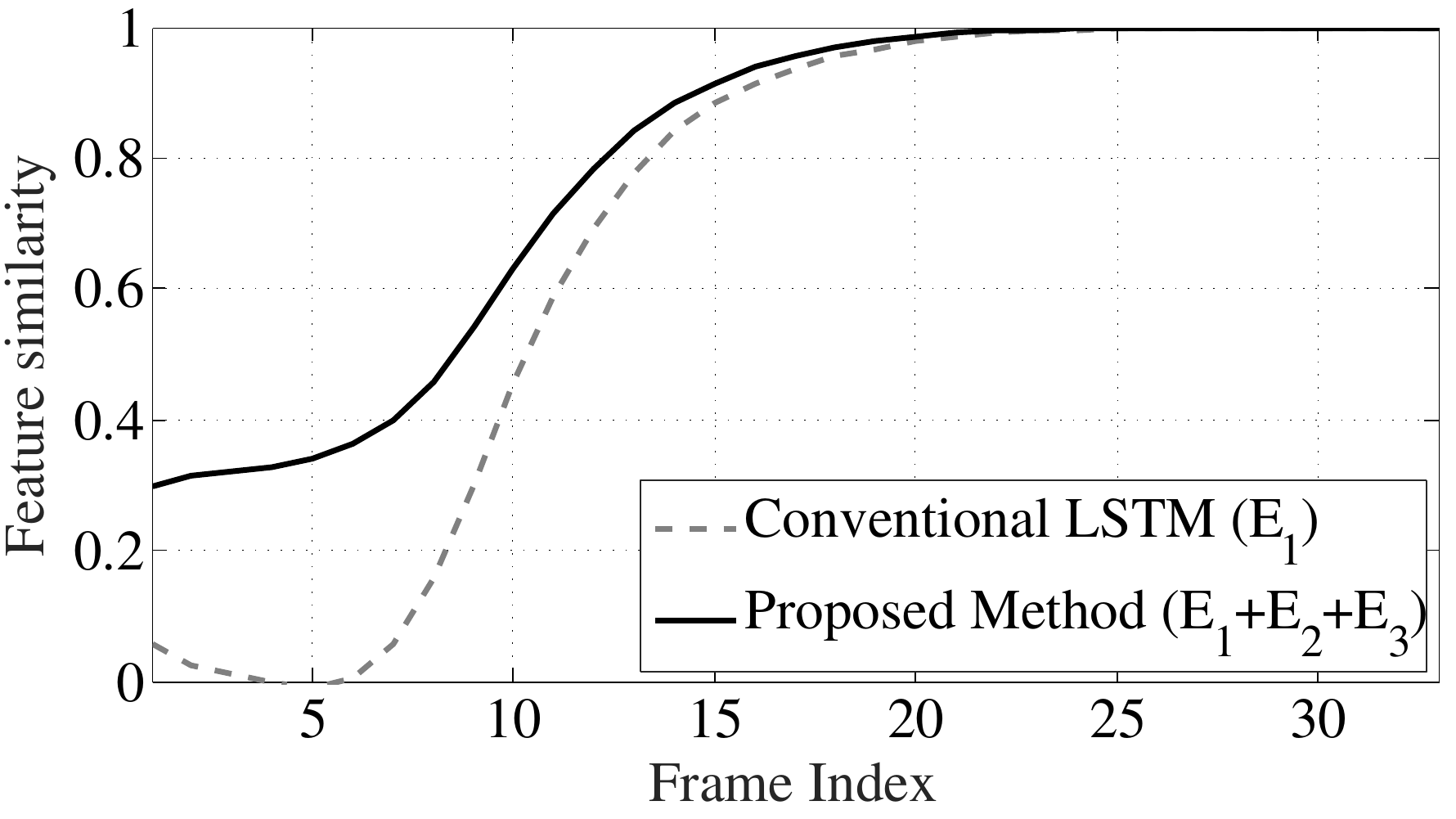}
             \caption{}
             \label{fig5:d}
     \end{subfigure}
     \begin{subfigure}[b]{0.3\textwidth}
             \centering
             \includegraphics[width=1\textwidth,height=2.5cm,keepaspectratio]{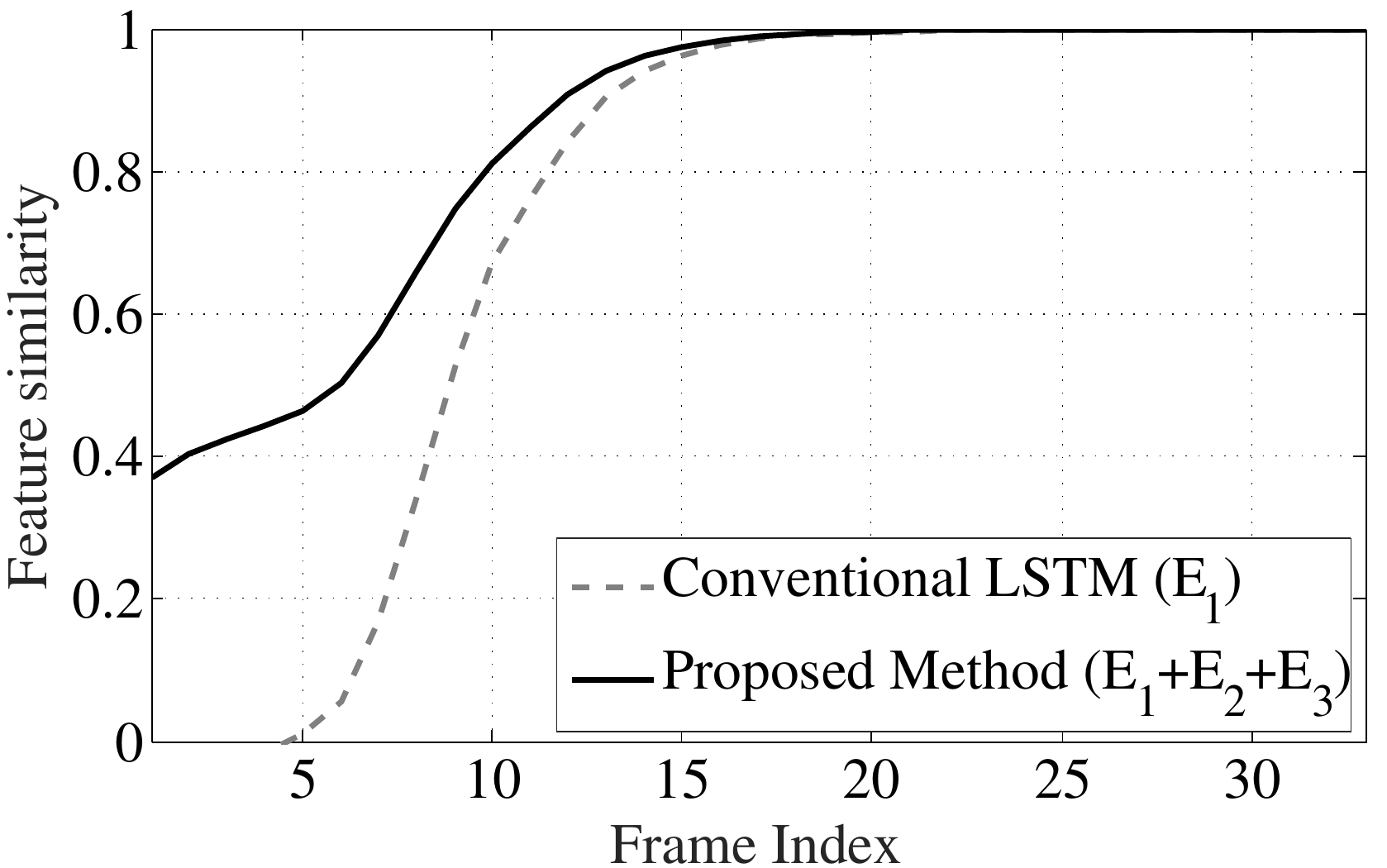}
             \caption{}
             \label{fig5:e}
      \end{subfigure}
      \begin{subfigure}[b]{0.3\textwidth}
             \centering
             \includegraphics[width=1\textwidth,height=2.5cm,keepaspectratio]{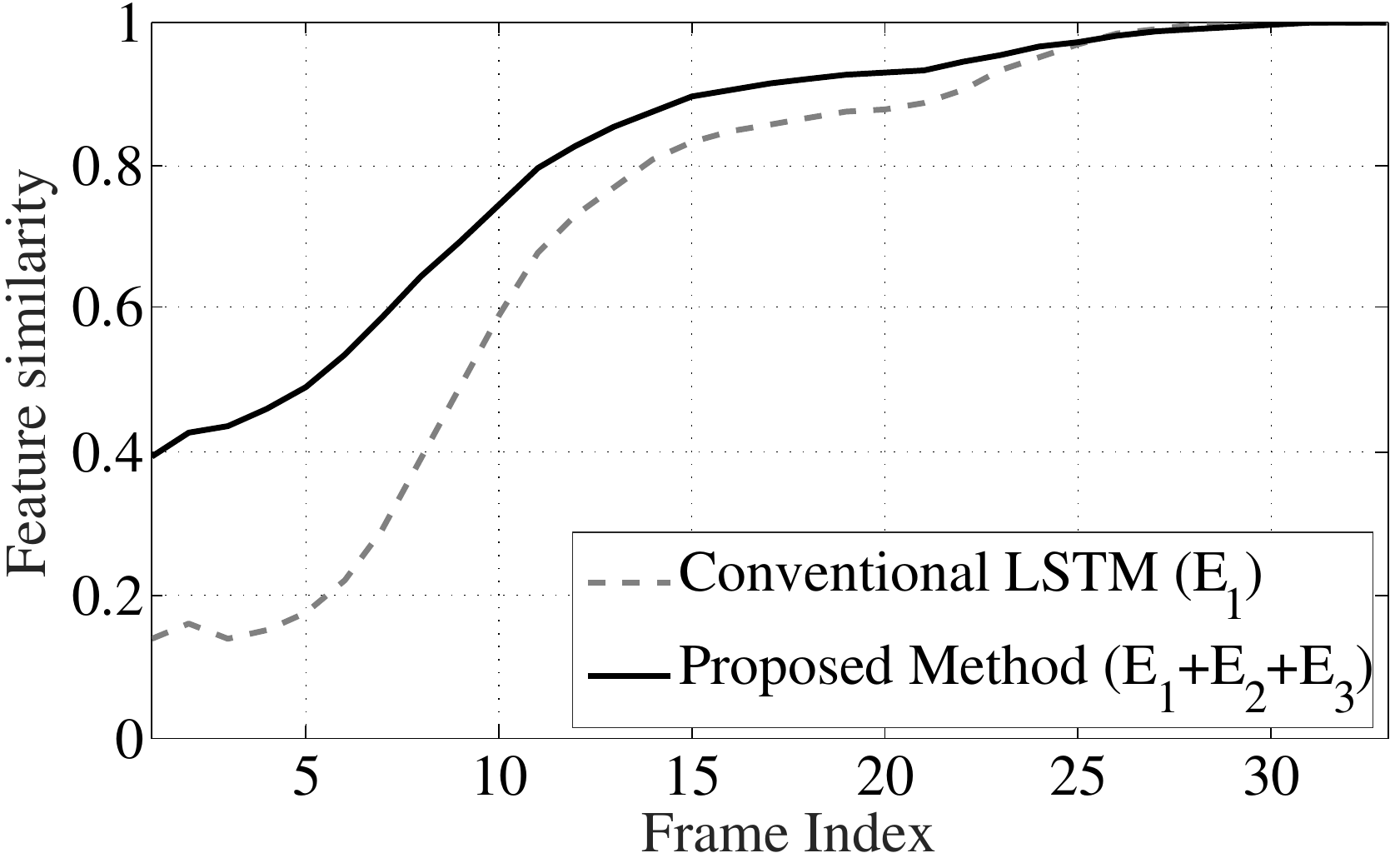}
             \caption{}
             \label{fig5:f}
      \end{subfigure}

\caption{Effect of the proposed method on the feature simiarity between intermediate partial sequence features and the full sequence features. The figure shows the average cosine similarity values between the obtained features at each frame on-the-fly (partial sequence inputs) and the apex frame features (full sequence). Each figure represents a different expression: (a) anger (b) disgust, (c) fear, (d) happiness, (e) sadness and (f) surprise.}
\label{fig:5}
\end{figure*}

\subsection{Evaluating the proposed method for on-the-fly prediction with partial sequence inputs}
In this experiment, we investigate the effectiveness of the proposed objective terms in improving the prediction on-the-fly with partial sequence inputs. For each sequence, prediction was performed at each frame using the learned spatio-temporal features of the partial expression sequence extending from the beginning of sequence until the frame in question). After the prediction was performed on all frames of the sequences (i.e., all partial sequences included in that sequence), recognition rate was calculated.

As previously mentioned, each expression sequence is labeled with one expression, but the sequence intensity changes from zero (at onset) to one at the apex frame. Hence, generating labels at the frame-level is necessary to evaluate the recognition rate at different expression intensities. To that end, the expression intensity was utilized to generate frame-level labels as shown in Figure~\ref{fig:3}. With a given expression intensity threshold (for example expression intensity = 0.2, as shown in the figure), all frames with expression intensity above that threshold are assumed as expressive images of that label (shown in green). On the other hand, all frames below that expression intensity threshold are considered as perceived natural frames (shown in pink). Accordingly, the recognition rate at that intensity threshold was calculated.

Figure~\ref{fig:4} shows a comparison between the obtained recognition rates for three different models. The first model was trained with a conventional LSTM (i.e. using the objective term $E_1$ only). The other two models were trained by incrementally adding the objective terms (i.e., one model was trained with $E_1 + E_2$, and $E_1 + E_2 + E_3$, respectively). As shown in the figure, the proposed objective terms significantly improve the prediction performance compared to the conventional LSTM method in terms of the recognition rate. More importantly, it can be seen that most of the improvement occurs in expression intensity range [0.3, 0.7].  These expression intensities can be considered subtle non-apex frames. It should also be noted that the prediction was only performed using a partial expression sequence (which extends from the beginning of the expression until that frame). This means that the proposed method can be readily applied to on-the-fly (interactive) applications. As such, the prediction can be performed after each frame is fed to the system, by considering the LSTM state of previous frames.

Notice the performance degradation at the expression intensity threshold = 1. At that point the prediction is performed with a full sequence. The number of expression sequences is significantly smaller than the number of frames, which justifies the resulting in the degradation in the prediction shown in the Figure\ref{fig:4}.
\subsection{Effect of the proposed method on the learned spatio-temporal features}
In this experiment, Oulu-CASIA dataset was utilized to analyze the effect of the proposed method on the learned features. To improve the prediction with partial sequences, the proposed objective function should increase the feature similarity between intermediate frame features (with partial sequence inputs) and the apex frame (with the full sequence). To validate that assumption, cosine similarity was calculated between the spatio-temporal features (L$_1$ output from Table~\ref{Table:1}) obtained from each frame and the frame obtained at the last frame (apex frame). Note that the features were obtained by feeding the LSTM the frames leading to the frame in question, to mimic the on-the-fly prediction process. The average of the features similarities from all the sequences of the same expression class are plotted in Figure~\ref{fig:5}. As the figure shows, the proposed method increases the features similarity between the intermediate frames and the apex frame compared to the conventional LSTM. This verifies that the learned model can predict the expression more accurately at earlier frames, alleviating the prediction delay problem.
\section{Conclusion}
In this paper, we proposed a new spatio-temporal feature learning method, which allow prediction with partial sequences on-the-fly. To achieve that, we proposed utilizing an estimated expression intensity to generate frame-level (dense) labels, which can be used to regulate the training of an LSTM with a novel objective function.  The purpose of the proposed objective function is to generate features that could robustly predict the expression with partial (incomplete) expression sequences, on-the-fly. This can solve the delay problem that could occur when using an LSTM on-the-fly prediction framework, with weakly labeled data (one label per expression sequence). Comprehensive experiments were conducted on two facial expression datasets. The experimental results showed that the proposed method improved the FER performance on sequence level prediction. The results also verified that the proposed method improved the prediction with partial expression sequence inputs at non-apex frames. These results indicate that the proposed method can solve the expression prediction delay problem and can be deployed in an on-the-fly scenario (i.e., interactive environments).
\section{Acknowledgment}
This work was supported by the Institute for Information \& communications Technology Promotion(IITP) grant funded by the Korea government(MSIT) (No. 2017-0-01778, Development of Explainable Human-level Deep Machine Learning Inference Framework)
\bibliography{refs}
\bibliographystyle{aaai}
\end{document}